\documentclass[fleqn,10pt]{wlscirep}
\usepackage[utf8]{inputenc}
\usepackage[T1]{fontenc}
\usepackage{algorithm, algpseudocode}
\usepackage{pgfplots}
\usepackage{caption}
\usepackage{subcaption}
\pgfplotsset{compat=1.17}
\title{Classification and regression of trajectories rendered as images via 2D Convolutional Neural Networks}

\author[1,$\dagger$]{Mariaclaudia Nicolai}
\author[1,2,$\dagger$]{Raffaella Fiamma Cabini}
\author[1,2,*]{Diego Ulisse Pizzagalli}
\affil[1]{Euler institute, Univeristà della Svizzera italiana, Lugano, Switzerland}
\affil[2]{International Center for Advanced Computing in Medicine (ICAM), University of Pavia, Pavia, Italy}

\affil[$\dagger$]{Contributed equally}
\affil[*]{Correspondence to pizzad@usi.ch}


\begin{abstract}
Trajectories can be regarded as time-series of coordinates, typically arising from motile objects. Methods for trajectory classification are particularly important to detect different movement patterns, while methods for regression to compute motility metrics and forecasting.

Recent advances in computer vision have facilitated the processing of trajectories rendered as images via artificial neural networks with 2d convolutional layers (CNNs). This approach leverages the capability of CNNs to learn spatial hierarchies of features from images, necessary to recognize complex shapes. Moreover, it overcomes the limitation of other machine learning methods that require input trajectories with a fixed number of points.

However, rendering trajectories as images can introduce poorly investigated artifacts such as information loss due to the plotting of coordinates on a discrete grid, and spectral changes due to line thickness and aliasing.

In this study, we investigate the effectiveness of CNNs for solving classification and regression problems from synthetic trajectories that have been rendered as images using different modalities. The parameters considered in this study include line thickness, image resolution, usage of motion history (color-coding of the temporal component) and anti-aliasing. Results highlight the importance of choosing an appropriate image resolution according to model depth and motion history in applications where movement direction is critical.

\end{abstract}

\begin{document}

\flushbottom
\maketitle
%
%
\thispagestyle{empty}

\section*{Introduction}
Trajectories arise from dynamic systems that evolve over time and are classically used to describe the sequence of positions that an object occupies while moving in space and time. This type of trajectories can be generated by various types of objects, such as planets, vehicles, people, cells, and particles, and are typically recorded by imaging devices or other position tracking systems. Analyzing the trajectories that objects follow is, therefore, crucial in a variety of scientific areas. For example, in astrophysics trajectory analysis allows the study of planet interaction \cite{zurlo2022orbital}, in traffic management tracking vehicles can help optimize flow and reduce congestion \cite{mazzarello2007traffic}. In life sciences, analyzing the movement of people can provide insight into their health and optimize sport performance \cite{barros2007analysis} while tracking the movement of cells observed via time-lapse microscopy can offer valuable information about their behavior and interactions with the microenvironment \cite{PizzagalliLatino2019, Beltman2009, Pizzagalli2018}.

Two primary tasks in trajectory analysis are regression and classification. Regression aims to predict continuous variables associated with the trajectory, such as motility metrics or forecasting future positions. Classification focuses on assigning discrete labels to the trajectory, such as the type of movement or the behavior that the object exhibited. Several methods can be employed to perform classification and regression of object trajectories. Traditional approaches often rely on the design of hand-made features and the use of statistical models such as linear regression or support vector machines to perform the prediction task~\cite{svensson2018untangling}. However, in recent years, Deep Learning (DL) models have gained popularity due to their ability to capture complex patterns in data. These models are particularly effective with high-dimensional or non-linear trajectory data~\cite{sen2019think, ju2020graph}.

DL methods for trajectory classification follow two distinct approaches: coordinate-based and image-based. Coordinate-based methods directly use the object’s sequence of spatial coordinates over time as input. The model can learn temporal dependencies and patterns from the raw trajectory data either considering them all at the same time, or employing recurrent architectures such as recurrent neural network and long-short term memory~\cite{cabini2024fast}. Image-based methods, on the other hand, involve transforming the trajectories into visual representations, which are then fed into DL models typically used for computer vision tasks, such as convolutional neural networks (CNNs). 

The image-based approach is particularly promising due to the increasing number and remarkable performances obtained by DL architectures for computer vision tasks that can be translated for trajectory analysis applications.
However, while there is a large literature body for coordinate-based methods \cite{ismail2019deep, ruiz2021great, abanda2019review}, image-based methods are less explored.

In this work, we will focus on image-based approaches for the analysis of 2D trajectories. Specifically, we will address both classification and regression tasks on trajectories of motile objects generated synthetically by varying motility parameters. The goal is to characterize how different line-rendering techniques used to convert time-series into visual representation affect the performance of DL models. By understanding these effects, we aim to provide insights into how to best utilize image-based methods for trajectory analysis.

\section*{Results}
\subsection*{An image dataset of trajectories rendered with different modalities}
To test 2D-CNNs for classification and regression tasks we constructed a dataset with $72000$ images representing 2D trajectories rendered with $36$ different rendering methods. Initially we generated 2000 trajectories using the algorithm described in \ref{alg}. Then we rendered them as images varying the following rendering parameters.

\begin{figure}[!h]
\centering
\includegraphics[width=\linewidth]{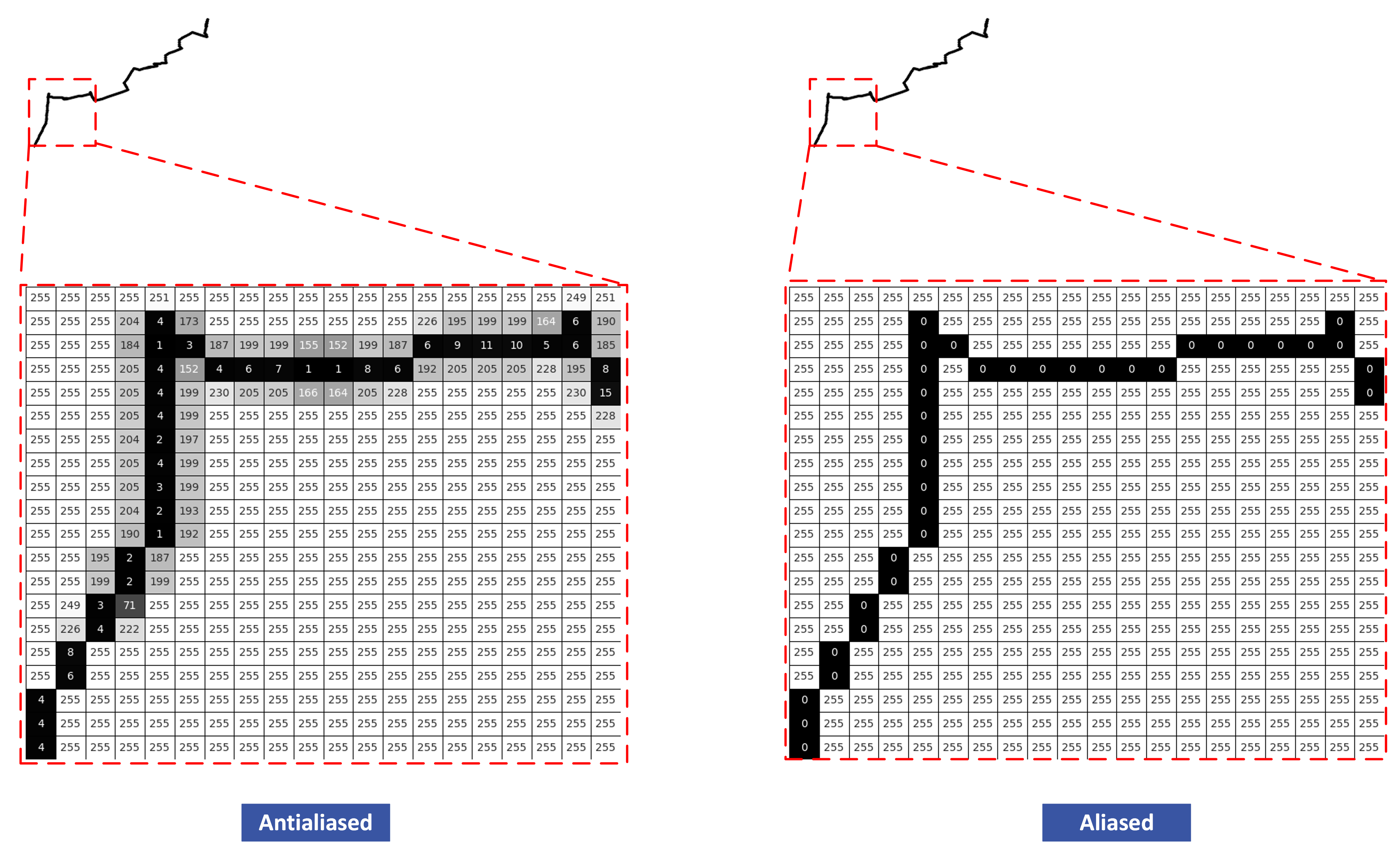}
\caption{Anti-aliased vs aliased image reading.}
\label{fig:aliasing}
\end{figure}

\begin{figure}[!h]
\centering
\includegraphics[width=\linewidth]{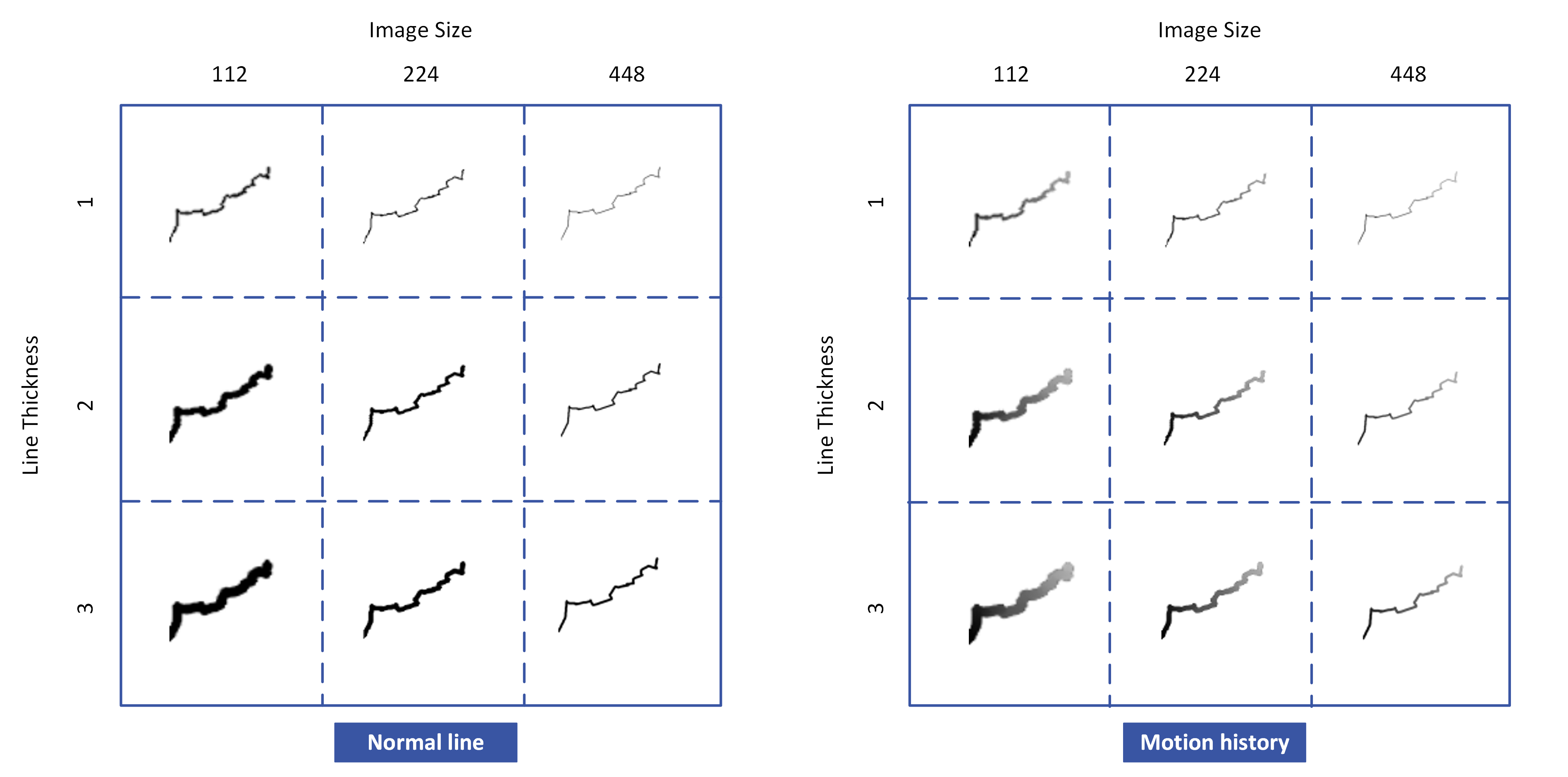}
\caption{Dataset of synthetic trajectories.}
\label{fig:data}
\end{figure}

\begin{enumerate}
    \item \emph{image resolution:} $112\times112$, $224\times224$, or $448\times448$ pixels;
    \item \emph{line thickness:} 1, 2, or 3 pixels;
    \item \emph{line color pattern:} trajectories were visualized using two distinct line patterns:
    \begin{itemize}
        \item the \emph{normal line pattern} represents the trajectory with a fixed color, displaying only the spatial information while losing the temporal aspect;
        \item the \emph{motion history line pattern} incorporates temporal information by varying the gray-level of the line along the trajectory: the initial point of the trajectory is represented with gray level of 0, this value increases for the successive points, reaching the maximum value of 255 at the final point of the trajectory. This gradient effect highlights both the spatial and temporal evolution of the trajectory;
    \end{itemize}
    \item \emph{aliasing effects:} we applied two effects, also depicted in Figure~\ref{fig:aliasing}:
    \begin{itemize}
        \item \emph{aliasing:} an under-sampling effect that occurs when representing a trajectory on a discrete grid, often leading to jagged or stair-step edges;
        \item \emph{anti-aliasing:} a technique used to reduce aliasing by smoothing jagged edges through the interpolation of additional pixels, resulting in smoother trajectories.
    \end{itemize}
\end{enumerate}
An overview of all the rendering modalities used in this study is presented in Figure~\ref{fig:data}.

\subsection*{Classification of trajectories from objects moving with different motion modes}
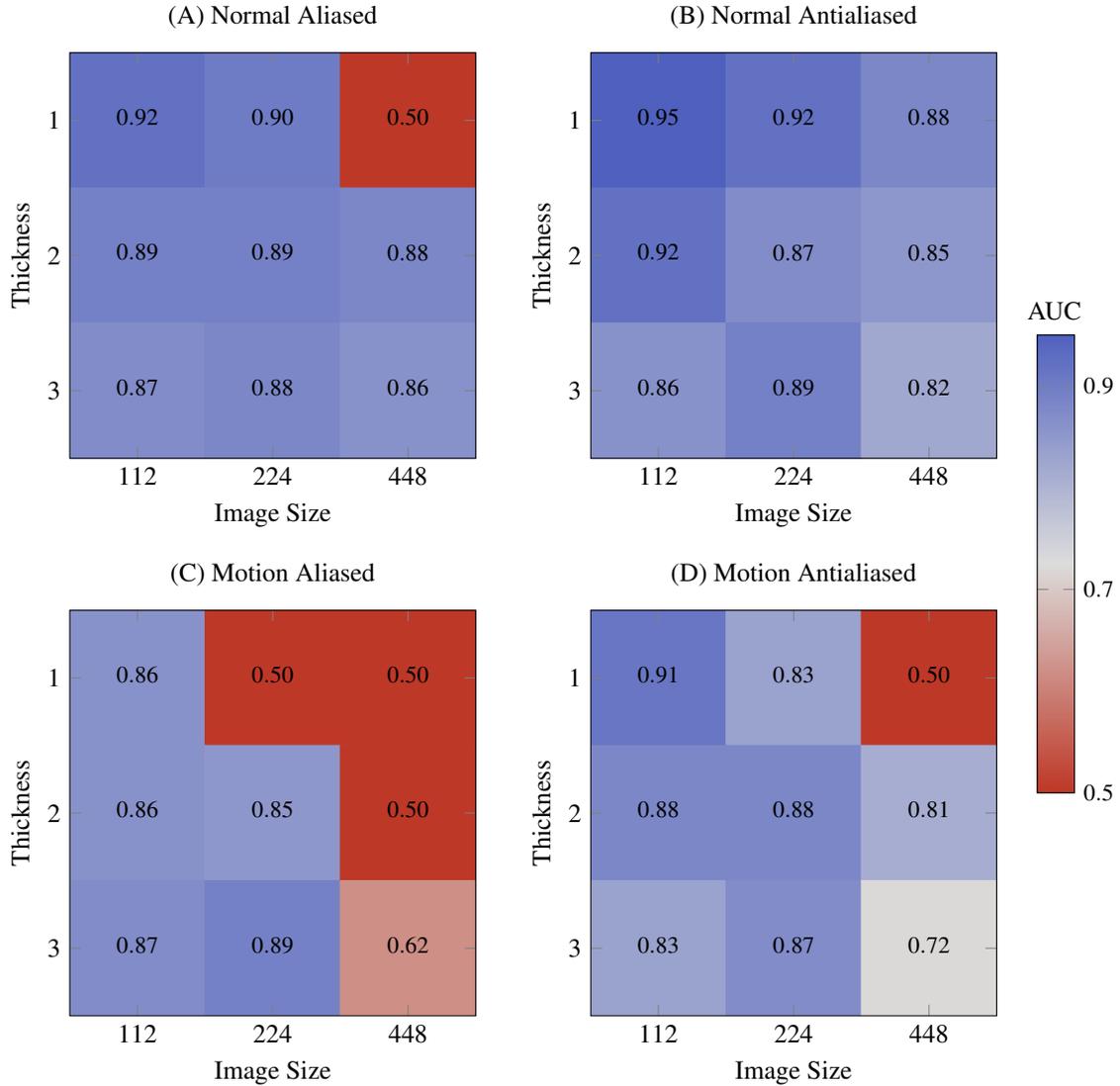
\begin{figure}[!h]
\centering
\pgfplotsset{
    colormap={pastel}{
        rgb255(0cm)=(190,56,39);  
        rgb255(1cm)=(220,220,220); 
        rgb255(2cm)=(79,96,190)
    }
}

\begin{tikzpicture}

\begin{axis}[
    at={(0,0)}, anchor=south west, 
    width=0.40\textwidth,
    height=0.40\textwidth,
    title={(A) Normal Aliased},
    xlabel={Image Size},
    ylabel={Thickness},
    xtick={1, 2, 3},
    xticklabels={112, 224, 448},
    ytick={1, 2, 3},
    yticklabels={1, 2, 3},
    colormap name=pastel,
    point meta min=0.5,
    point meta max=0.95,
    nodes near coords,
    every node near coord/.append style={font=\small, yshift=-1.2ex, /pgf/number format/.cd, fixed, precision=2, fixed zerofill},
    enlargelimits=false
]
\addplot[
    matrix plot,
    mesh/rows=3,
    mesh/cols=3,
    point meta=explicit
] table [meta=C] {
    x y C
    1 1 0.92
    2 1 0.90
    3 1 0.50
    1 2 0.89
    2 2 0.89
    3 2 0.88
    1 3 0.87
    2 3 0.88
    3 3 0.86
};
\end{axis}

\begin{axis}[
    at={(7cm,0)}, anchor=south west,
    width=0.40\textwidth,
    height=0.40\textwidth,
    title={(B) Normal Antialiased},
    xlabel={Image Size},
    ylabel={Thickness},
    xtick={1, 2, 3},
    xticklabels={112, 224, 448},
    ytick={1, 2, 3},
    yticklabels={1, 2, 3},
    colormap name=pastel,
    point meta min=0.5,
    point meta max=0.95,
    nodes near coords,
    every node near coord/.append style={font=\small, yshift=-1.2ex, /pgf/number format/.cd, fixed, precision=2, fixed zerofill},
    enlargelimits=false
]
\addplot[
    matrix plot,
    mesh/rows=3,
    mesh/cols=3,
    point meta=explicit
] table [meta=C] {
    x y C
    1 1 0.95
    2 1 0.92
    3 1 0.88
    1 2 0.92
    2 2 0.87
    3 2 0.85
    1 3 0.86
    2 3 0.89
    3 3 0.82
};
\end{axis}

\begin{axis}[
    at={(0,-7.5cm)}, anchor=south west,
    width=0.40\textwidth,
    height=0.40\textwidth,
    title={(C) Motion Aliased},
    xlabel={Image Size},
    ylabel={Thickness},
    xtick={1, 2, 3},
    xticklabels={112, 224, 448},
    ytick={1, 2, 3},
    yticklabels={1, 2, 3},
    colormap name=pastel,
    point meta min=0.5,
    point meta max=0.95,
    nodes near coords,
    every node near coord/.append style={font=\small, yshift=-1.2ex, /pgf/number format/.cd, fixed, precision=2, fixed zerofill},
    enlargelimits=false
]
\addplot[
    matrix plot,
    mesh/rows=3,
    mesh/cols=3,
    point meta=explicit
] table [meta=C] {
    x y C
    1 1 0.86
    2 1 0.50
    3 1 0.50
    1 2 0.86
    2 2 0.85
    3 2 0.50
    1 3 0.87
    2 3 0.89
    3 3 0.62
};
\end{axis}

\begin{axis}[
    at={(7cm,-7.5cm)}, anchor=south west,
    width=0.40\textwidth,
    height=0.40\textwidth,
    title={(D) Motion Antialiased},
    xlabel={Image Size},
    ylabel={Thickness},
    xtick={1, 2, 3},
    xticklabels={112, 224, 448},
    ytick={1, 2, 3},
    yticklabels={1, 2, 3},
    colormap name=pastel,
    point meta min=0.5,
    point meta max=0.95,
    nodes near coords,
    every node near coord/.append style={font=\small, yshift=-1.2ex, /pgf/number format/.cd, fixed, precision=2, fixed zerofill},
    enlargelimits=false
]
\addplot[
    matrix plot,
    mesh/rows=3,
    mesh/cols=3,
    point meta=explicit
] table [meta=C] {
    x y C
    1 1 0.91
    2 1 0.83
    3 1 0.50
    1 2 0.88
    2 2 0.88
    3 2 0.81
    1 3 0.83
    2 3 0.87
    3 3 0.72
};
\end{axis}

\begin{axis}[
    at={(12.5cm,-4.5cm)}, anchor=south west, 
    hide axis,
    scale only axis,
    height=0.35\textwidth,
    width=0.4cm,
    colormap name=pastel,
    colorbar,
    point meta min=0.5,
    point meta max=0.95,
    colorbar style={
        ytick={0.5, 0.70, 0.90},
        yticklabel style={font=\small},
        ylabel={AUC},
        ylabel style={yshift=1.4cm, xshift=3.4cm, rotate=-90}, 
        y label style={yshift=0cm}, 
    }
]
\addplot [draw=none] coordinates {(0,0)};
\end{axis}

\end{tikzpicture} 
\caption{Heatmaps of median AUC values obtained by the classification CNN on the test set across the three independent training. Each heatmap shows AUC values across varying image sizes ($112\times112$, $224\times224$, $448\times448$ pixels) and line thicknesses ($1$, $2$, $3$ pixels). Panel A and B represent normal line pattern (Normal), with aliasing (left) and anti-aliasing (right) effects applied, while panels C and D correspond to the motion history line pattern (Motion), with aliasing (left) and anti-aliasing (right). Blue colors indicate higher AUC values, while red colors represent lower AUCs.}
\label{fig:classification}
\end{figure}

Figure~\ref{fig:classification} represents the median area under the receiver operating characteristic curve (AUC) values for the classification task with different rendering conditions. The results represent the mean AUC values from three independent training runs. The AUC values are displayed for combinations of line thickness ($y$-axis) and image size ($x$-axis). Each heatmap represents a specific condition obtained by the combination of line pattern (normal line and motion history) and aliasing effects (aliasing and anti-aliasing).

Panel A shows the AUC values for normal lines with aliasing effect applied. The highest AUC ($0.92$) is achieved with the smallest image size ($112\times112$ pixels) and the thinnest line thickness (thickness $1$). However, as the image size increases to $448\times448$ pixels, the performance drops, particularly for thickness $1$, where the AUC falls to $0.50$. The classifier maintains relatively stable performance across the other image sizes and line thicknesses, with AUC values ranging from $0.86$ to $0.89$.

Panel B depicts the results for normal lines with anti-aliasing effect. In this case, the best AUC value is equal to $0.95$ for the smallest image size and line thickness of $1$, which is a slightly better AUC value compared to the aliased version. As the image size increases, the AUC decreases but not as drastically as in the aliased condition. For larger image sizes, the AUC stabilizes between $0.82$ and $0.92$, suggesting that anti-aliasing effect helps maintain a higher level of performance even with increasing image size.

In panel C, AUC values for motion history lines with aliasing effect are displayed. The 2D-CNN's classification performance is more unstable in this condition. For larger image sizes ($224\times224$ and $448\times448$ pixels), the AUC decreases, particularly for line thickness equal to $1$ (AUC equal to $0.50$). In contrast, with smaller image sizes ($112\times112$ pixels), the AUC remains relatively high, ranging form  $0.86$ to $0.87$ across all the line thicknesses, indicating that smaller image sizes are less affected by the aliasing effect with motion history lines.

Panel D shows the performance with motion history lines and anti-aliasing effect. The highest AUC value ($0.91$) is observed with the smallest image size and line thickness of $1$, which is slightly lower value than the normal line anti-aliased case. As the image size increases, the AUC tends to decrease, reaching a minimum of $0.50$ for the largest image size and smallest line thickness. However, the AUC stabilizes between $0.83$ and $0.91$ for larger line thicknesses and all the image sizes, illustrating that anti-aliasing effect helps mitigate performance decrease with motion history lines observed for larger image sizes.

Overall, the figure demonstrates that smaller image sizes ($112\times112$ pixels) yield the highest AUC values, especially when anti-aliasing is applied. The application of the anti-aliasing effect consistently improves classifier performance, particularly for larger image sizes, where aliasing causes significant drops in AUC values. Motion history line pattern generally reduces the AUC values compared to normal lines, with performance drop most visible for larger image sizes in motion history aliased conditions. This highlights the importance of anti-aliasing in reducing the negative impact of the motion history pattern and of the aliasing effect on classification performance.

\subsection*{Regression of trajectory directionality}
\begin{figure}[!h]
\centering
\pgfplotsset{
    colormap={pastel}{
        rgb255(0cm)=(79,96,190);
        rgb255(1cm)=(220,220,220);
        rgb255(2cm)=(190,56,39)
    }
}

\begin{tikzpicture}

\begin{axis}[
    at={(0,0)}, anchor=south west, 
    width=0.40\textwidth,
    height=0.40\textwidth,
    title={(A) Normal Aliased},
    xlabel={Image Size},
    ylabel={Thickness},
    xtick={1, 2, 3},
    xticklabels={112, 224, 448},
    ytick={1, 2, 3},
    yticklabels={1, 2, 3},
    colormap name=pastel,
    point meta min=0.04,
    point meta max=0.10,
    nodes near coords,
    every node near coord/.append style={font=\small, yshift=-1.2ex,/pgf/number format/.cd, fixed, precision=2, fixed zerofill},
    enlargelimits=false
]
\addplot[
    matrix plot,
    mesh/rows=3,
    mesh/cols=3,
    point meta=explicit
] table [meta=C] {
    x y C
    1 1 0.07
    2 1 0.07
    3 1 0.07
    1 2 0.07
    2 2 0.06
    3 2 0.07
    1 3 0.07
    2 3 0.07
    3 3 0.07
};
\end{axis}

\begin{axis}[
    at={(7cm,0)}, anchor=south west,
    width=0.40\textwidth,
    height=0.40\textwidth,
    title={(B) Normal Antialiased},
    xlabel={Image Size},
    ylabel={Thickness},
    xtick={1, 2, 3},
    xticklabels={112, 224, 448},
    ytick={1, 2, 3},
    yticklabels={1, 2, 3},
    colormap name=pastel,
    point meta min=0.04,
    point meta max=0.10,
    nodes near coords,
    every node near coord/.append style={font=\small, yshift=-1.2ex,/pgf/number format/.cd, fixed, precision=2, fixed zerofill},
    enlargelimits=false
]
\addplot[
    matrix plot,
    mesh/rows=3,
    mesh/cols=3,
    point meta=explicit
] table [meta=C] {
    x y C
    1 1 0.07
    2 1 0.07
    3 1 0.07
    1 2 0.07
    2 2 0.07
    3 2 0.07
    1 3 0.06
    2 3 0.07
    3 3 0.06
};
\end{axis}

\begin{axis}[
    at={(0,-7.5cm)}, anchor=south west,
    width=0.40\textwidth,
    height=0.40\textwidth,
    title={(C) Motion Aliased},
    xlabel={Image Size},
    ylabel={Thickness},
    xtick={1, 2, 3},
    xticklabels={112, 224, 448},
    ytick={1, 2, 3},
    yticklabels={1, 2, 3},
    colormap name=pastel,
    point meta min=0.5,
    point meta max=0.95,
    point meta min=0.04,
    point meta max=0.10,
    nodes near coords,
    every node near coord/.append style={font=\small, yshift=-1.2ex,/pgf/number format/.cd, fixed, precision=2, fixed zerofill},
    enlargelimits=false
]
\addplot[
    matrix plot,
    mesh/rows=3,
    mesh/cols=3,
    point meta=explicit
] table [meta=C] {
    x y C
    1 1 0.07
    2 1 0.07
    3 1 0.08
    1 2 0.05
    2 2 0.06
    3 2 0.10
    1 3 0.04
    2 3 0.05
    3 3 0.07
};
\end{axis}

\begin{axis}[
    at={(7cm,-7.5cm)}, anchor=south west,
    width=0.40\textwidth,
    height=0.40\textwidth,
    title={(D) Motion Antialiased},
    xlabel={Image Size},
    ylabel={Thickness},
    xtick={1, 2, 3},
    xticklabels={112, 224, 448},
    ytick={1, 2, 3},
    yticklabels={1, 2, 3},
    colormap name=pastel,
    point meta min=0.04,
    point meta max=0.10,
    nodes near coords,
    every node near coord/.append style={font=\small, yshift=-1.2ex, /pgf/number format/.cd, fixed, precision=2, fixed zerofill},
    enlargelimits=false
]
\addplot[
    matrix plot,
    mesh/rows=3,
    mesh/cols=3,
    point meta=explicit
] table [meta=C] {
    x y C
    1 1 0.07
    2 1 0.07
    3 1 0.08
    1 2 0.04
    2 2 0.05
    3 2 0.08
    1 3 0.05
    2 3 0.05
    3 3 0.08
};
\end{axis}

\begin{axis}[
    at={(12.5cm,-4.5cm)}, anchor=south west, 
    hide axis,
    scale only axis,
    height=0.35\textwidth,
    width=0.4cm,
    colormap name=pastel,
    colorbar,
    point meta min=0.04,
    point meta max=0.10,
    colorbar style={
        ytick={0.04, 0.07, 0.10},
        yticklabel style={font=\small},
        ylabel={MAE},
        ylabel style={yshift=2cm, xshift=3.4cm, rotate=-90},
        y label style={yshift=0cm}, 
    }
]
\addplot [draw=none] coordinates {(0,0)};
\end{axis}

\end{tikzpicture} 
\caption{Heatmaps of median MAE values obtained by the regression CNN on the test set across the three independent training. Each heatmap shows MAE values across varying image sizes ($112\times112$, $224\times224$, $448\times448$ pixels) and line thicknesses ($1$, $2$, $3$ pixels). Panel A and B represent normal line pattern (Normal), with aliasing (left) and anti-aliasing (right) effects applied, while panels C and D correspond to the motion history line pattern (Motion), with aliasing (left) and anti-aliasing (right). Red colors indicate higher MAE values, while blue colors represent lower MAE values.}
\label{fig:regression}
\end{figure}
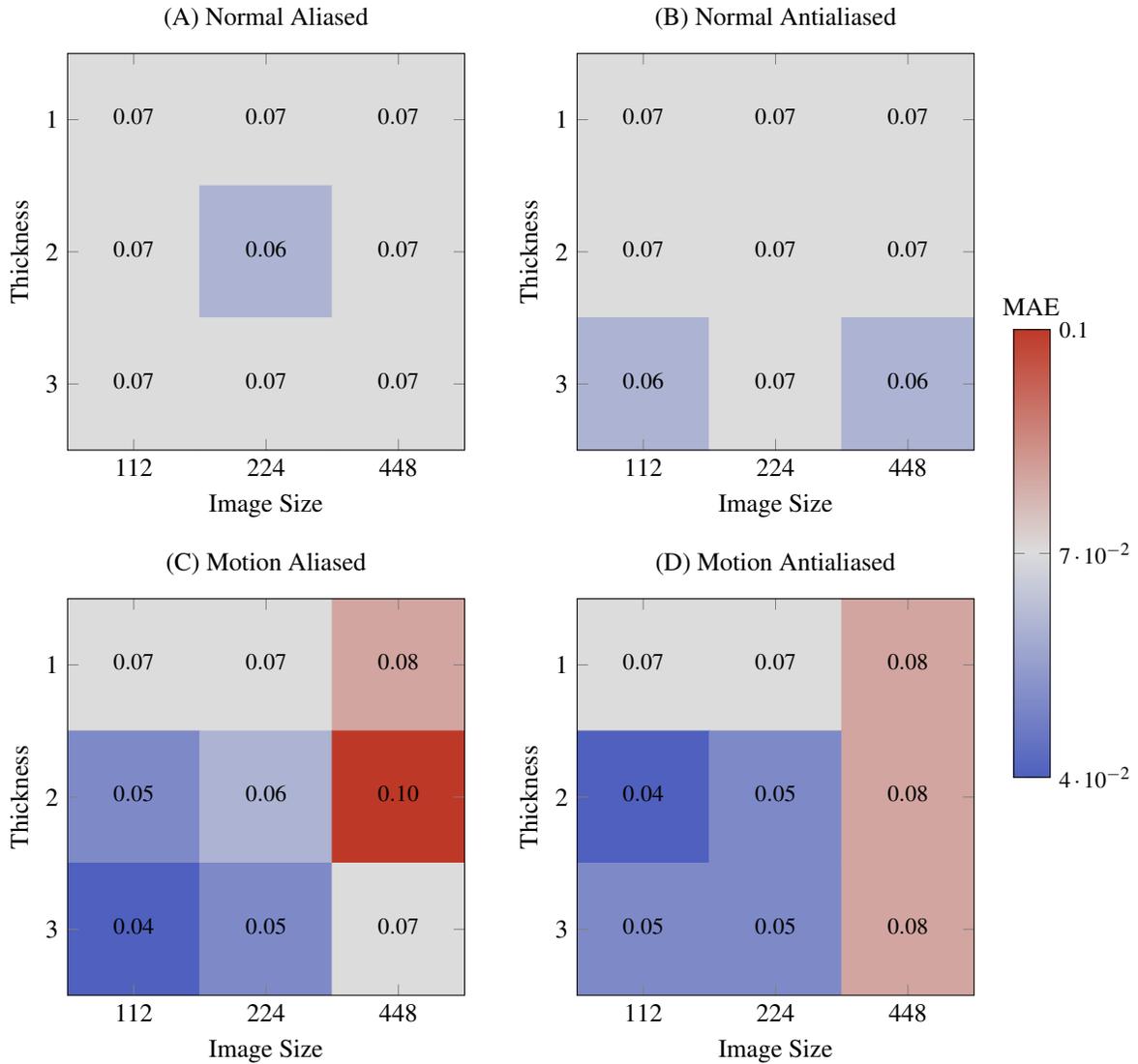

Figure~\ref{fig:regression} represents the median mean absolute error (MAE) values for the regression task with different rendering conditions. The results represent the mean MAE values from three independent training runs. The MAE values are displayed for combinations of line thickness ($y$-axis) and image size ($x$-axis). Each heatmap represents a specific condition obtained by the combination of line pattern (normal line and motion history) and aliasing effects (aliasing and anti-aliasing).

Panel A represents the normal line aliased condition. MAE values are mostly uniform across all image sizes and thicknesses, varying around $0.07$. The uniformity of the results implies that image size and thickness have little impact on the error in this condition.

Panel B shows the results obtained with the application of the anti-aliasing effect to the normal line pattern. MAEs remain similar to the previous condition, with values consistently around $0.07$ across most combinations of line thickness and image size. Compared to the aliased condition, the application of anti-aliasing does not introduce significant variations in error, and the performance remains stable across all thicknesses and image sizes.

Panel C, which represents the motion history line pattern with aliasing effect, shows greater variability in the MAE compared to normal lines. For bigger line thicknesses (thickness $3$) and smaller image sizes ($112\times112$ pixels), the MAE decreases significantly, reaching $0.04$. However, as the image size increases or the line thickness decreases, the MAE increases. For the largest image size, the error ranges between $0.07$ and $0.10$. This suggests that the motion history pattern combined with the aliasing effect has an impact on MAE.

Panel D, which shows results for motion history pattern with anti-aliasing effect, presents a similar pattern to the motion history aliased case but with slightly reduced MAE values for some conditions (e.g. for image size equal to $112\times112$ pixels and line thickness of $2$, MAE equal to $0.04$). However, for larger image sizes ($448\times448$ pixels), the MAE remains around $0.08$, regardless of the line thickness.

Overall, the figure demonstrates that MAE remains relatively stable with all the normal line conditions, regardless of whether aliasing or anti-aliasing are applied. However, when the motion history pattern is used, especially in the presence of aliasing, the error increases for larger image sizes and thinner line thicknesses. On the other hand, in both the motion history conditions, with smaller image sizes and thicker line thicknesses, the lowest error values are achieved ($0.04$) than those observed with the normal lines. Anti-aliasing helps reduce the MAE particularly for smaller image sizes and thicker line thicknesses, but the error still increases for larger images. This indicates that both motion history and aliasing effect affect the 2D-CNN's performance, although motion history pattern can, in some cases, result in lower error values compared to normal conditions.

\section*{Discussion}
In this study, we explored the use of CNNs to classify and regress 2D trajectories represented as images.
Despite being described since the 1980s \cite{lecun1989}, only during the last decade the use of CNNs has grown significantly, due to their ability to learn spatial hierarchies of features in images, becoming nowadays a fundamental block of most computer vision applications.

For this work, we employed a simple 2D-CNN architecture similar to ConvNet \cite{liu2022}, which is one of the most widely used and easily implemented CNN architectures. This architecture was chosen to establish a baseline performance for trajectory classification and regression. However, it is important to note that the results obtained with this architecture may differ from those using deeper or more complex architectures. For instance, increasing the number of layers and parameters could potentially improve performances, particularly when working with input images with higher resolution.

Amongst the rendering methods, we evaluated the use of motion history images, a technique that color-codes points along the trajectory based on their temporal coordinate. Motion history could potentially preserve temporal dynamics when the rendered trajectories intersect. Our findings show that motion history did not improve classification accuracy for the directional memory task, suggesting that for this specific task, the added temporal dimension might not have been essential. However, when applied to the regression task for predicting the directionality of trajectories, motion history led to improvements, confirming the importance of the temporal dimension in tasks where the direction of movement must be taken into account.

Our studies were performed on bidimensional trajectories having two spatial coordinates evolving over time. To process trajectories having a single coordinate with 2D-CNNs  projection into a higher-dimensional space can be considered \cite{saad2007}. Conversely, to process trajectories having multiple variables, dimensionality reduction techniques to map the data into a lower-dimensional space can be used.

Overall, the study demonstrates the potential of CNNs for trajectory analysis, but also highlights unobvious effects of rendering methods that might impact on performances if not considered properly.

\section*{Methods}

\subsection*{Synthetic trajectory generator}
The synthetic trajectory generator was designed to simulate the position of a moving object over time within a two-dimensional plane. The simulator employs a motility model accounting for stochastic changes in direction and directional memory. Let define the position of an object at time step $t$ with $x$ and  $y$ for the horizontal and vertical coordinates respectively.
The position of the object at time step $t+1$ is determined from its position at time step $t$ the rules of Algorithm~\ref{alg}.

\begin{algorithm}[h]
\caption{Synthetic trajectory generator}
\label{alg}
\begin{algorithmic}[1]
\State Given $N$ cells initialized in $(x^n_0 , y^n_0) \gets (0,0)$ with $n = 1, \dots, N$;
\State Sample $\eta_x \sim \mathcal{N}(0, \sigma)$ and $\eta_y \sim \mathcal{N}(0, \sigma)$
\State Initialize  $(x_1^n,y_1^n) \gets (\eta_x,\eta_y)$
\For{each time step $t = 2, 3, \ldots, T$}
    \State Sample $\alpha \sim \text{Uniform}(0, 1)$
    \If{$\alpha  < p$}
        \State $\Delta x^n \gets x_t^n - x_{t-1}^n$
        \State $\Delta y^n \gets y_t^n - y_{t-1}^n$
    \Else
        \State  Sample $\eta_x \sim \mathcal{N}(0, \sigma)$ and $\eta_y \sim \mathcal{N}(0, \sigma)$
        \State $\Delta x^n \gets \eta_x$
        \State $\Delta y^n \gets \eta_y$
    \EndIf
    \State Compute the data change:
    \State $x_{t+1}^n \gets x_t^n + \Delta x^n$
    \State $y_{t+1}^n \gets y_t^n + \Delta y^n$
\EndFor
\end{algorithmic}
\end{algorithm}

The model describes a movement characterized by two processes:
\begin{itemize}
    \item \textit{persistence of movement:} with probability $p$, the object tends to maintain a constant direction and speed, simulating a uniform rectilinear motion.
    \item \textit{random movement:} with probability $1-p$, the object experiences a sudden change in its direction and speed, introducing an element of randomness into the motion.
\end{itemize}
The combination of these two processes allows the generation of synthetic trajectories that can simulate the behavior of a cell during migration: the cell tends to maintain the direction of movement with probability $p$ and changes direction randomly with probability $1-p$. 

In Algorithm~\ref{alg}, $\eta_x$ and $\eta_y$ are two independent random variables sampled from a normal distribution $\mathcal{N}(0, \sigma)$ centered at $0$ with a standard deviation of $\sigma$. $\alpha$ is a random variable sampled from a uniform distribution $\text{Uniform}(0, 1)$ over the interval $[0, 1]$. The model is characterized by four free parameters: the probability $p$, which we will refer to as \textit{directional memory}, the variance $\sigma$ of the normal distribution, the total number of time steps $T$ and the total number of cells $N$. In the experiments of this study, we kept $T=100$, $N=1000$ and $\sigma=0.1$ fixed, while we explored different values of the directional memory.

\subsection*{Datasets for classification and regression}
The classification dataset comprised 2000 trajectories with two different classes: 1000 trajectories of class A generated with a directional memory value of $0.9$, and 1000 trajectories of class B with a directional memory value of $0.7$.

For the regression task, the dataset also consisted of 2000 samples. The target variable for the regression analysis was the directionality of the trajectories, which was calculated for each trajectory using the procedure described in the following paragraph. 

Both datasets were divided into training, validation, and testing subsets, with 1600 samples designated for the training set and 200 samples each for the validation and testing sets.

\subsubsection*{Computation of directionality}
For each cell trajectory we quantified its directionality as follows. Let $l$  be the total path length followed by the object from time $t = 1, \dots, T$ and let $d$ be the displacement, i.e., the length of the vector connecting the first with the last point. Then, the directionality $D$ is defined as the ratio between displacement and total path length, converging to 1 for straight trajectories and decreasing for non-linear trajectories:
\begin{equation}
    D = \frac{d}{l} = \frac{\sqrt{(x_T - x_1)^2 + (y_T - y_1)^2}}{\sum_{t=1}^{T-1} \sqrt{(x_{t+1} - x_t)^2 + (y_{t+1} - y_t)^2}}
\end{equation}

Directionality values less than one indicate that the trajectory is not linear or straight. Exceptional cases in which the centroid of the object did not move had $l = 0$. In these cases directionality was set to zero.

\subsubsection*{Data preprocessing}
Before rendering the images, two preprocessing steps were applied to the sequence of coordinates for each trajectory. First, the initial points of each trajectory were centered in the image by translating the coordinates so that each trajectory starts from the central coordinates of the image matrix. This translation was achieved by subtracting the first coordinate value from all subsequent coordinates. Next, a normalization step was conducted to confine the trajectory within a square frame. This involved identifying the maximum absolute value among the trajectory coordinates and dividing all coordinates by this maximum value. Finally, the processed trajectories were visualized by plotting the $x$ and $y$ coordinates using different rendering modalities.

We utilized Python 3.8.10 and OpenCV 4.9.0.80 to process trajectory data and generate images with various rendering modalities. 


\subsection*{Deep Leaning model}
The 2D-CNN model was built using the TensorFlow Python library. It comprises a sequence of four convolutional layers with ReLU activations and $3\times3$ kernel size, designed to extract meaningful features from input images. All the convolutional layers contain $32$ filters, except for the third layer, which has $64$ filters. After each convolutional layer, a MaxPooling layer with a $2\times2$ pool size is applied to reduce the spatial dimensions of each feature map.

After the last pooling layer, the output is flattened into a one-dimensional array and passed through two fully connected layers. The first fully connected layer contains 32 units with a ReLU activation function. For classification tasks, the final fully connected layer has $2$ units with a sigmoid activation function. For regression tasks, this layer has a single unit with a linear activation function to enable continuous output predictions.

\begin{figure}[!h]
\centering
\includegraphics[width=0.8\linewidth]{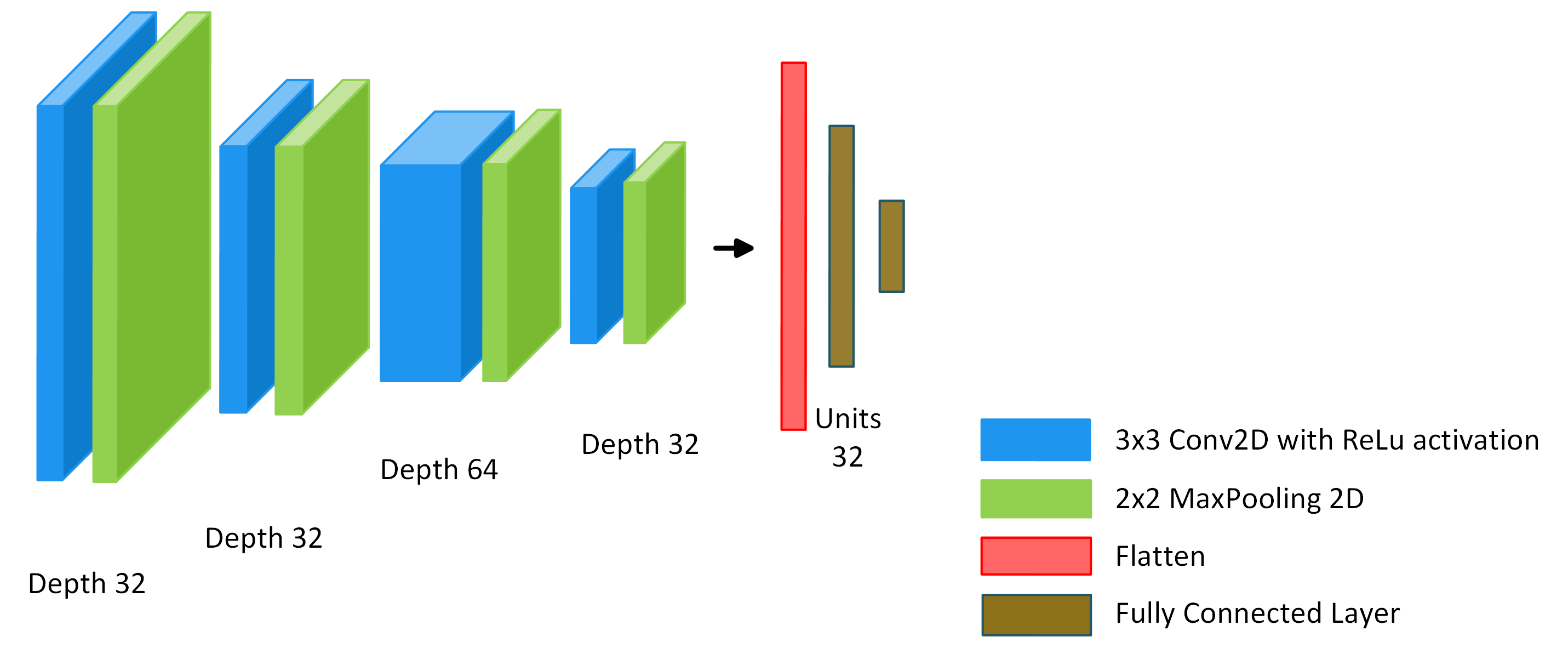}
\caption{2D-CNN model used for classification and regression tasks.}
\label{fig:cnn}
\end{figure}

Figure~\ref{fig:cnn} illustrates the architecture of the 2D-CNN.

\subsection*{Training and evaluation strategies}
Both the classification and regression 2D-CNNs were trained, validated, and tested on three random and disjoint partitions of the dataset. The training of both the models was conducted separately for 60 epochs, minimizing the cross-entropy loss for classification and the mean squared error for regression.  The Adam optimizer was used with a learning rate of $0.001$. Accuracy and MAE were used as additional metrics to monitor the training process for classification and regression, respectively. The 2D-CNN's weights that achieved the best loss during training were selected.

To evaluate how different rendering methods affected 2D-CNN performance, each model was trained over three independent training, and the median of the evaluation metrics on the test set was reported as the final performance. For classification, AUC was calculated, while for regression, we used the MAE.

\bibliography{sample}



\section*{Acknowledgements}
This work has been supported financially by the FIR grant (Fondo Istituzionale per la Ricerca, USI), to DUP

\section*{Author contributions statement}
MN performed experiments, developed software and and wrote the manuscript. RFC supervised the work, developed software and wrote the manuscript. DUP conceptualized the project, supervised the work and wrote the manuscript.

\section*{Additional information}
Authors declare no competing interests.

\newpage
\section*{Figures}

\end{document}